\documentclass[english]{article}
\usepackage[T1]{fontenc}
\usepackage[latin9]{inputenc}
\usepackage{babel}
\usepackage{refstyle}
\usepackage{url}
\usepackage{amsbsy}
\usepackage{amssymb}
\usepackage{graphicx}
\usepackage{esint}
\usepackage[unicode=true,
 bookmarks=true,bookmarksnumbered=false,bookmarksopen=false,
 breaklinks=false,pdfborder={0 0 1},backref=false,colorlinks=false]
 {hyperref}
\hypersetup{pdftitle={Normalizing flows for novelty detection},
 pdfauthor={Maximilian Schmidt, Marko Simic},
 pdfsubject={Invertible Flows, Normalizing Flows, Novelty detection}}

\makeatletter

\AtBeginDocument{\providecommand\figref[1]{\ref{fig:#1}}}
\RS@ifundefined{subsecref}
  {\newref{subsec}{name = \RSsectxt}}
  {}
\RS@ifundefined{thmref}
  {\def\RSthmtxt{theorem~}\newref{thm}{name = \RSthmtxt}}
  {}
\RS@ifundefined{lemref}
  {\def\RSlemtxt{lemma~}\newref{lem}{name = \RSlemtxt}}
  {}

\usepackage{hyperref}
\usepackage[accepted]{icml2019}

\renewcommand\figref{\Figref}

\makeatother

\begin{document}
\icmltitlerunning{Normalizing flows for novelty detection}

\twocolumn[
\icmltitle{Normalizing flows for novelty detection in industrial time series data}
\begin{icmlauthorlist}
\icmlauthor{Maximilian Schmidt}{to}
\icmlauthor{Marko Simic}{to}
\end{icmlauthorlist}

\icmlaffiliation{to}{Ascent Robotics Inc., Japan}

\icmlcorrespondingauthor{Maximilian Schmidt}{maximilian@ascent.ai}

\icmlkeywords{Invertible Flows, Normalizing Flows, Novelty Detection}

\vskip 0.3in
]

\printAffiliationsAndNotice{}  
\begin{abstract}
Flow-based deep generative models learn data distributions by transforming
a simple base distribution into a complex distribution via a set of
invertible transformations. Due to the invertibility, such models
can score unseen data samples by computing their exact likelihood
under the learned distribution. This makes flow-based models a perfect
tool for novelty detection, an anomaly detection technique where unseen
data samples are classified as normal or abnormal by scoring them
against a learned model of normal data. We show that normalizing flows
can be used as novelty detectors in time series. Two flow-based models,
Masked Autoregressive Flows and Free-form Jacobian of Reversible Dynamics
restricted by autoregressive MADE networks, are tested on synthetic
data and motor current data from an industrial machine and achieve
good results, outperforming a conventional novelty detection method,
the Local Outlier Factor.
\end{abstract}

\section{Introduction}

Novelty detection comprises anomaly detection algorithms which first
learn a model of the normal data, and at inference time compute a
novelty score of unseen data samples under the learned model. Data
samples are flagged as normal or abnormal by comparing their novelty
score to a learned decision boundary. In that sense, novelty detection
can be seen as a one-class classification task. One important characteristic,
and appealing property, of novelty detection algorithms is that during
training time they do not require access to instances of anomalous
data. This is important in situations in which we have little or no
anomalous data, and it is hard and costly to obtain such data. Because
of this, novelty detection is widely used in many domains, such as
medical diagnostics \citep{tarassenko1995novelty}, security of electronic
systems \citep{Patcha_2007}, or mobile robotics \citep{hornung2014model}.

\citet{pimentel2014review} classify novelty detection algorithms
in several groups: probabilistic, distance-based, reconstruction-based,
domain-based, and information-theoretic based techniques. For our
application, we initially chose four methods that belong to different
algorithm families outlined above: support vector machine (SVM) \citep{scholkopf2000support},
isolation forest (IF) \citep{liu2012isolation}, local outlier factor
detector (LOF) \citep{markus2000lof} as representative of conventional
machine learning algorithms, and normalizing flows (NF) \citep{kingma2016improved}
as representative of deep learning algorithms. After preliminary tests,
out of three mentioned conventional methods, LOF yielded best classification
accuracy, so we focused our further effort on LOF and normalizing
flows.

Normalizing flows are a class of deep generative models that leverage
invertible neural networks to learn a mapping between a simple base
distribution and a given data distribution. The invertibility allows
for two important use cases: generation of new data and classification
of input data samples by computing the likelihood of such samples.
The latter one makes them a suitable candidate algorithm for novelty
detection. Compared to classical algorithms, normalizing flows allow
us to flexibly constrain the learned distributions, for instance by
enforcing autoregressive property when modeling time series.

In this work, we demonstrate the applicability of normalizing flows
for novelty detection in time series. We apply two different flow
models, masked autoregressive flows (MAF) \citep{papamakarios2017masked}
and FFJORD \citep{Grathwohl2019} restricted by a Masked Autoencoder
for Distribution Estimation (MADE) architecture \citep{germain2015made}
to synthetic data and motor current time series data from an industrial
machine. Both flow-based models achieve superior results over the
local outlier factor method. Furthermore, we demonstrate the generation
of new data samples with the learned flow models which can give rise
to further use cases in the domain of anomaly detection and defect
analysis in industrial machines.

\section{Datasets\label{sec:Datasets}}

We create synthetic data to test the ability of the models for novelty
detection by generating random time series with a defined autocorrelation
function. We specify the autocorrelation function as $f(\Delta t)=\exp\left(-|\Delta t|/\tau\right)\cos\left(\Delta t/15\right)$
which defines the covariance matrix $\Sigma$ for a given length of
time series $T$ and compute its Cholesky decomposition $B$ such
that $\Sigma=BB^{\mathrm{T}}$. We then generate data samples by drawing
white noise samples $y\sim N(0,1)\in\mathbb{R}^{T}$and transforming
them to samples of our time series with given autocorrelation as $z=yB$.
We define normal samples to have $\tau=50$ and vary the decay time
to create abnormal samples. By construction, the inter-sample mean
at each timestep is 0. To keep the variance across samples at every
time step equal between normal and abnormal samples, we divide abnormal
samples $y^{\prime}$ by their inter-sample variance and multiply
by the variance of the normal time series: $y_{i}^{\prime}(t)\leftarrow y_{i}^{\prime}(t)/\mathrm{Var}_{j}(y_{j}^{\prime}(t))\cdot\mathrm{Var}{}_{j}(y_{j}(t))$.

To test on real data, we used a dataset that was generously provided
by Kawasaki Heavy Industries Ltd (KHI). Data is an electric motor
current signal (measured in Amperes), collected from the electrical
motor of one of KHI's products. Dataset contains both the signal during
motor's normal operation, and anomalous signal after a gearbox connected
to the motor experienced an undefined problem in its operation. There
are eight different patterns of motor operation, each pattern lasting
between 5 and 30 seconds. Signals were sampled with frequency of 500
Hz. Ratio of normal to anomalous data in the dataset was roughly equal,
but for model training purposes only a subset of normal data was used,
while held-out normal data and anomalous data were used for testing
purposes. One sample of normal and anomalous data is shown in \figref{Fig3_real_data}A.
In this work, we use a subset of one of eight patterns that, based
on classfication performance of the LOF, we found to be the most challenging
pattern for anomaly detection.

\section{Model setup}

We train three different models on our data: Masked Autoregressive
Flows, Continuous Normalizing Flows using Free-form Jacobian of Reversible
Dynamics with a MADE network, and Local Outlier Factor.
\begin{itemize}
\item MAF uses a stack of fixed number of affine layers whose scale and
shift parameters are computed by an autoregressive network, here implemented
using the MADE architecture \citep{germain2015made}. Given a latent
random variable $z_{t}\sim\pi(z_{t}),$ the transformed variable $x_{t}$
is computed as $x_{t}=z_{t}\odot\mu(\boldsymbol{x}_{1:t-1})+\sigma(\boldsymbol{x}_{1:t-1})\sim p(x_{t}|\boldsymbol{x}_{1:t-1})$,
where the scale and shift terms $\boldsymbol{\mu},\boldsymbol{\sigma}$
are efficiently computed by one forward pass through a MADE network.
We chose MAF over Inverse Autoregressive Flows (IAF) \citep{kingma2016improved}
because MAF offers fast evaluation of data likelihood which is essential
in novelty detection. IAF, on the other hand, offers fast generation
of new data but slow evaluation of test data. In our experiments we
chose the standard normal distribution as the base distribution $\boldsymbol{z}\sim N(0,I)$
and stack 5 coupling layers each with MADE networks consisting of
3 hidden layers with 256 units each and tanh activation function.
\item The FFJORD model \citep{Grathwohl2019} extends the idea of continuous
normalizing flows (CNF) \citep{chen2018neural} by an improved estimator
of the log-density of samples. CNF models the latent variable $\boldsymbol{z}$
with an ordinary differential equation $\mathrm{\partial}\boldsymbol{z}/\mathrm{\partial}\tilde{t}=f(\boldsymbol{z},\tilde{t})$
so that transforming from latent to data space is equivalent to integrating
the ODE from pseudo times $\tilde{t}_{0}$ to $\tilde{t}_{1}$: $\boldsymbol{x}=\boldsymbol{z}(t_{1})=\int_{t_{0}}^{t_{1}}f(\boldsymbol{z},t)\mathrm{d}t$
(see \citet{Grathwohl2019} for details). The function $f$ is represented
by a neural network. We here chose a MADE architecture to enforce
the autoregressive property between different time samples: $f(z_{t},\tilde{t})=f(z_{1:t-1},\tilde{t})$.
In our experiments, we use 2 hidden layers with 256 neurons and tanh
activation function. More details on the experiments in Section \ref{sec:Supplement}.
\item Local outlier factor is a distance based novelty detection algorithm
that assigns a degree of being an outlier to each data point. This
degree, local outlier factor, is determined by comparing the local
density of a data point to the local density of its neighboring points.
A point that has significantly lower local density than its neighbors
is considered to be an outlier. The main parameter used to influence
the algorithm's performance is \emph{MinPts} which specifies the number
of points to be considered as neighborhood of a datapoint $p$. \citet{markus2000lof}
establishes that inliers have value of LOF equal to 1, while LOF for
outliers is greater than 1. It also provides the tightness of the
lower and upper bound for outlier's LOF values.
\end{itemize}

\section{Experiments}

\begin{figure}
\begin{centering}
\includegraphics[clip,width=1\linewidth]{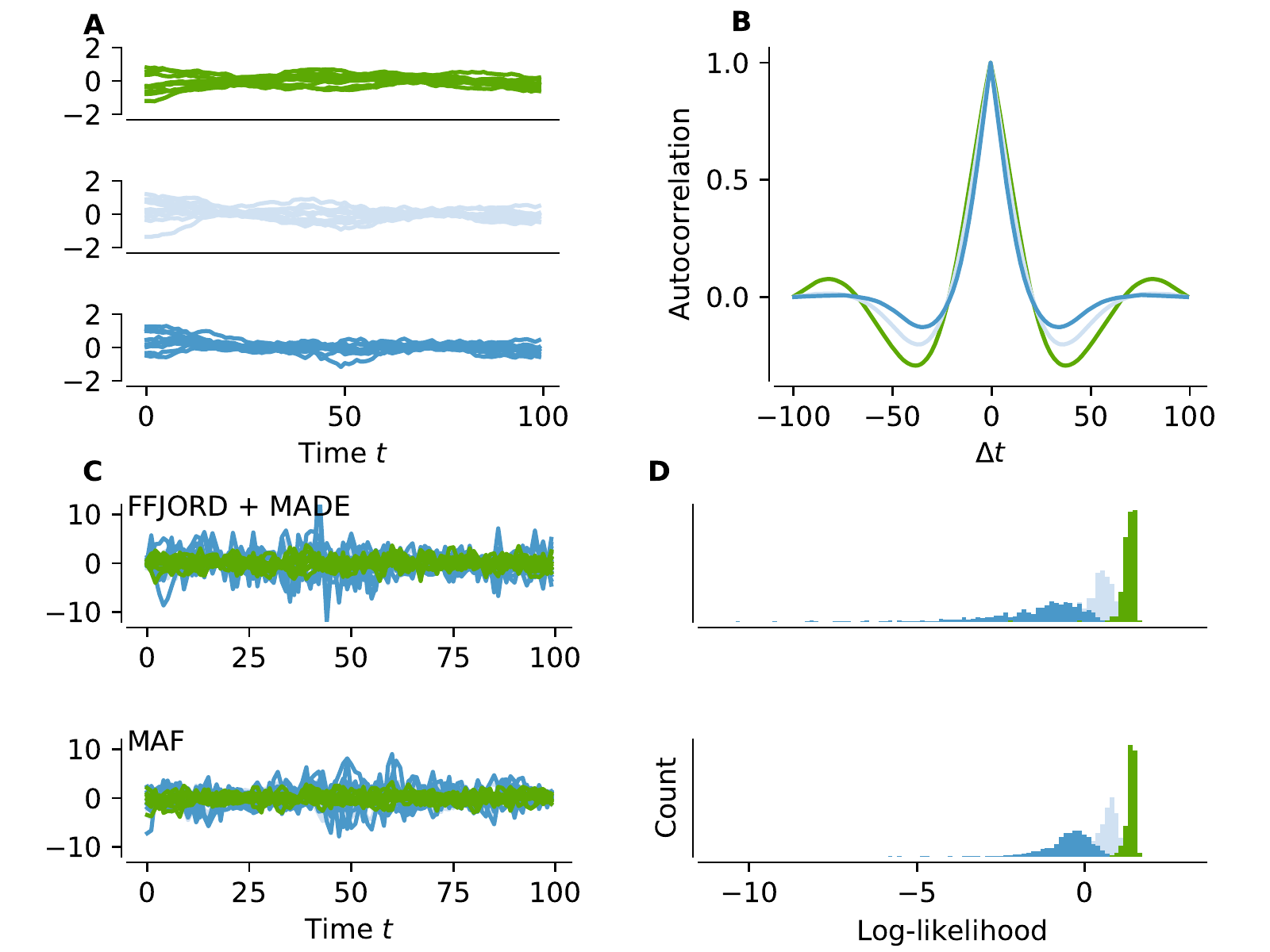}
\par\end{centering}
\caption{Flow models applied to synthetic data: normal data with $\tau=50$
(green) and abnormal data ($\tau=30$, light blue and $\tau=20$,
dark blue). A: 10 random samples of the time series. B: Mean autocorrelation
across 1000 samples. C: 10 samples transformed by applying FFJORD+MADE
and MAF. D: Histogram over mean likelihood values per time points
of normal and abnormal samples.}
\label{fig:Fig1_toy_data}
\end{figure}

We test our idea of using flow models as novelty detectors for time
series on a set of synthetic data where we control the deviation between
normal and abnormal samples. The data is created as time series with
a defined autocorrelation where we systematically vary the time constant
of the autocorrelation (see Section \ref{sec:Datasets} for details).
Normal and abnormal data only vary in their correlation between time
points, while the ensemble statistics (in terms of mean and variance
across samples) are equal at every time step (\figref{Fig1_toy_data}A,
B). This makes anomaly detection challenging because the model cannot
resort to simply representing mean and variance of time steps independently
but rather needs to learn a joint distribution representing temporal
correlations.

We train the two flow models on the normal data and then apply them
to unseen normal samples as well as abnormal data samples (\figref{Fig1_toy_data}C).
Both models transform the normal samples to approx. white noise while
the transformed abnormal samples clearly deviate from white noise.
Consistent with the visual inspection of transformed samples, the
models assign lower likelihood to abnormal samples than normal samples,
and this difference becomes more clear with increasing deviation of
abnormal samples, i.e. decreasing time constant of the autocorrelation
function (\figref{Fig1_toy_data}, dark blue vs. light blue in data
points).

\begin{figure}
\begin{centering}
\includegraphics[clip,width=1\linewidth]{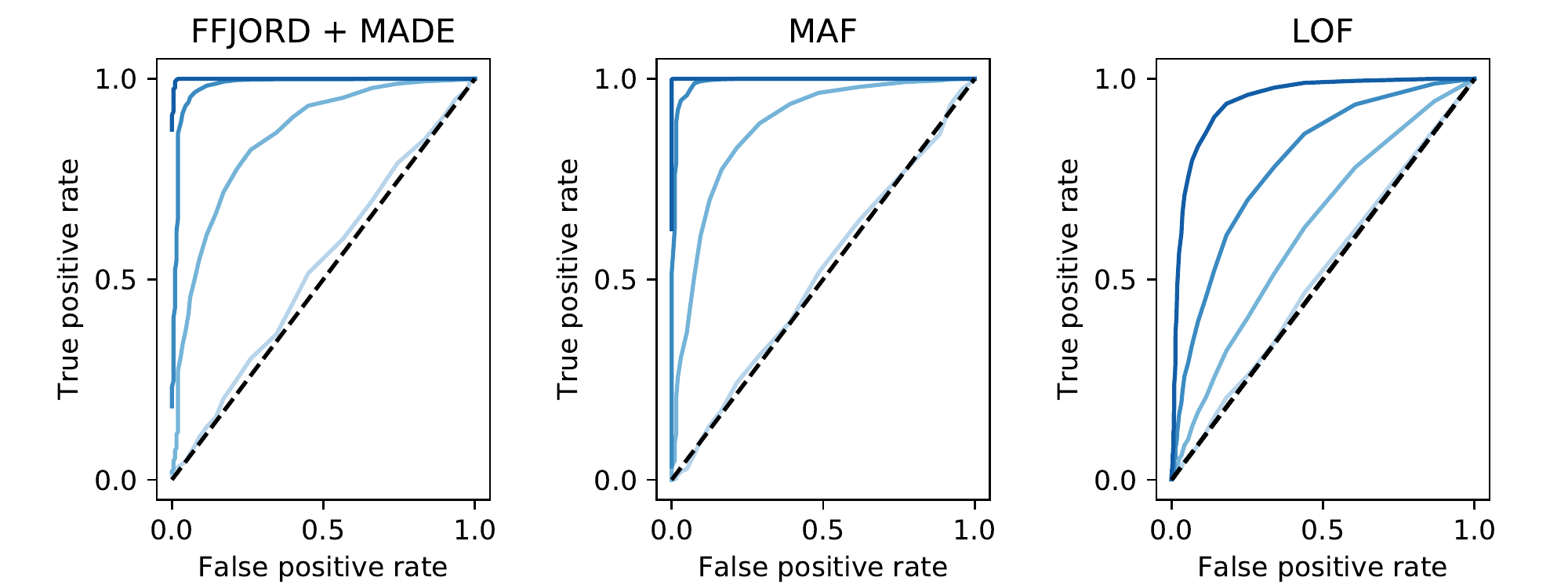}
\par\end{centering}
\caption{Novelty detection results for synthetic data. Receiver operating characteristic
(ROC) curves for all three tested models on abnormal synthetic data
with decreasing time constants from 50 (light blue) to 20 (dark blue).
The dashed line indicates chance level.}
\label{fig:Fig2_novelty_det}
\end{figure}

To use the trained model as a novelty detector, we define a decision
boundary, i.e. a likelihood value which separates abnormal from normal
data samples. To judge the quality of a novelty detector, a typical
metric is the receiver operating characteristic (ROC) curve. Given
a model and abnormal data samples, we vary the decision boundary and
measure the rate of false positive (normal data classified as abnormal)
versus true positive (abnormal data classified as abnormal). Thus,
the steeper the slope of the ROC curve, the better. For varying time
constants in the autocorrelation of the synthetic data, we compute
the ROC curve of the two flow-based models and compare them to the
local outlier factor (LOF) (\figref{Fig2_novelty_det}). Both flow-based
models quickly deviate from the chance-level at $\tau=50$ (in this
case, the test data is drawn from the same distribution as normal
data) for decreasing time constant. The LOF model does not impose
autoregressive constraints on the time series and thus, it performs
worse in modelling the temporal correlations between data points in
the time series.

\begin{figure}
\begin{centering}
\includegraphics[clip,width=1\linewidth]{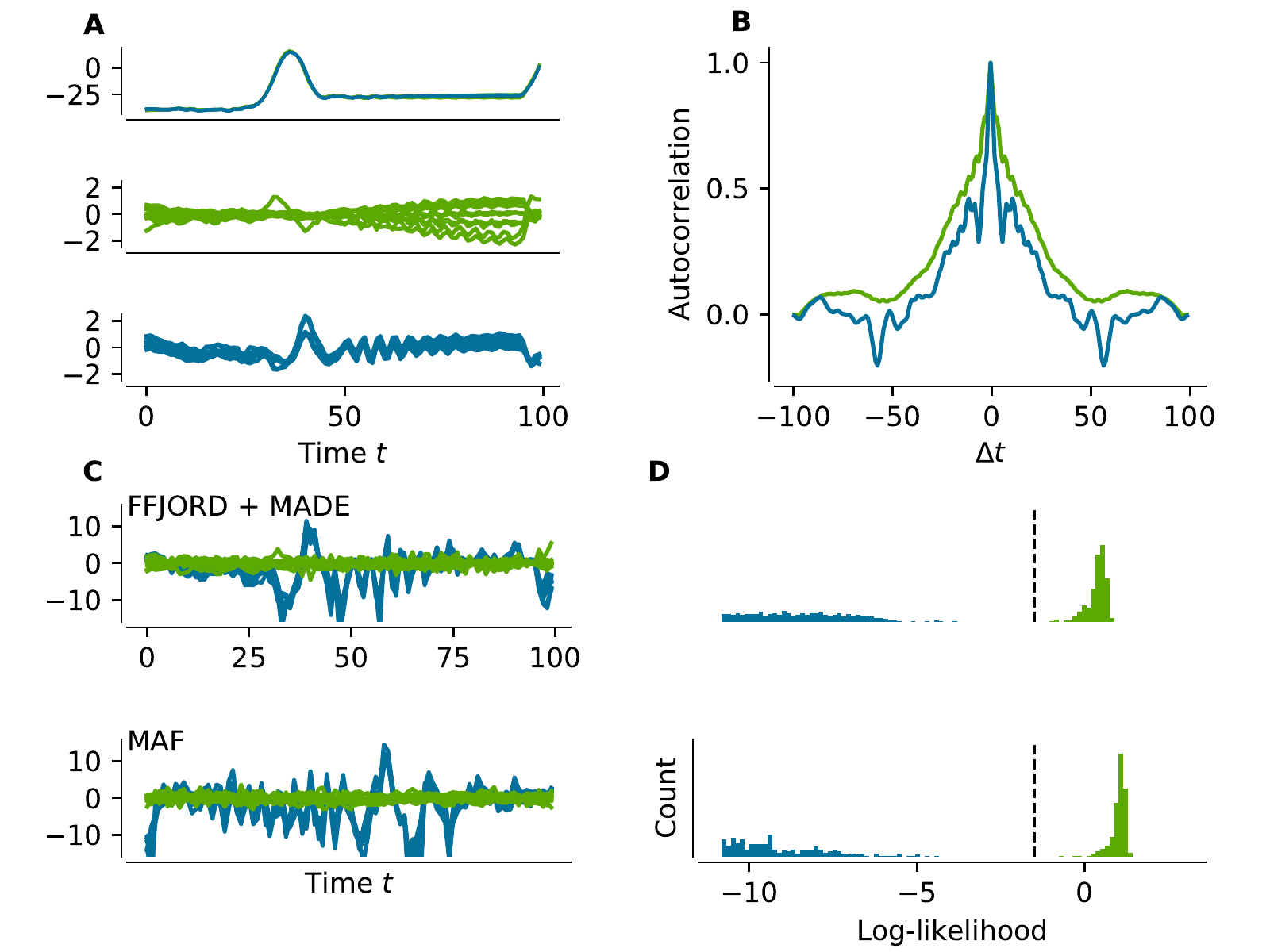}
\par\end{centering}
\caption{Flow models applied to real data: normal data (green) and abnormal
data (dark blue). A: 10 samples of normal and abnormal data in the
original domain (top) and centered with zero mean (middle and bottom).
B: Mean autocorrelation across samples. C: Transformed samples by
applying FFJORD+MADE and MAF. D: Histogram over likelihood values
of normal and abnormal samples. The vertical dashed line indicates
a possible decision boundary.}
\label{fig:Fig3_real_data}
\end{figure}

We test the feasibility of the flow-based models on real data (see
Section \ref{sec:Datasets} for details). Both normal and abnormal
time series follow a very similar global curve (\figref{Fig3_real_data}A,
top) and are visually hardly distinguishable. We thus split normal
samples into train and test data and subtract the mean across training
samples from normal and abnormal samples (\figref{Fig3_real_data}A,
middle and bottom). Abnormal and normal samples display different
temporal correlations (\figref{Fig3_real_data}B), and consequently
after training on the normal data both models clearly separate normal
test data and abnormal data (\figref{Fig3_real_data}C). We find that
the FFJORD+MADE diverges when applied to abnormal data samples after
the training has been performed for sufficient number of epochs (depending
on the hyperparameters), where very low training and test loss has
been reached. The ordinary differential equation becomes unstable
for samples deviating from the training distribution. For the purposes
of visualization, we thus decide to stop the training after 140 epochs
for \figref{Fig3_real_data}. While the normal samples are approximately
mapped to uncorrelated white noise, the transformed abnormal samples
clearly deviate from white noise. To score the samples, we again compute
the likelihood per time point and bin them into histograms (\figref{Fig3_real_data}D).
It is trivial to define a decision boundary which reaches $0\,\%$
false positive rate and $100\,\%$ true positive rate. The LOF model
reaches an accuracy of around $100\,\%$ true positive rate and $0.5\,\%$
false positive rate on this task. The flow-based models marginally
outperform the simpler method.

\begin{figure}
\begin{centering}
\includegraphics[clip,width=1\linewidth]{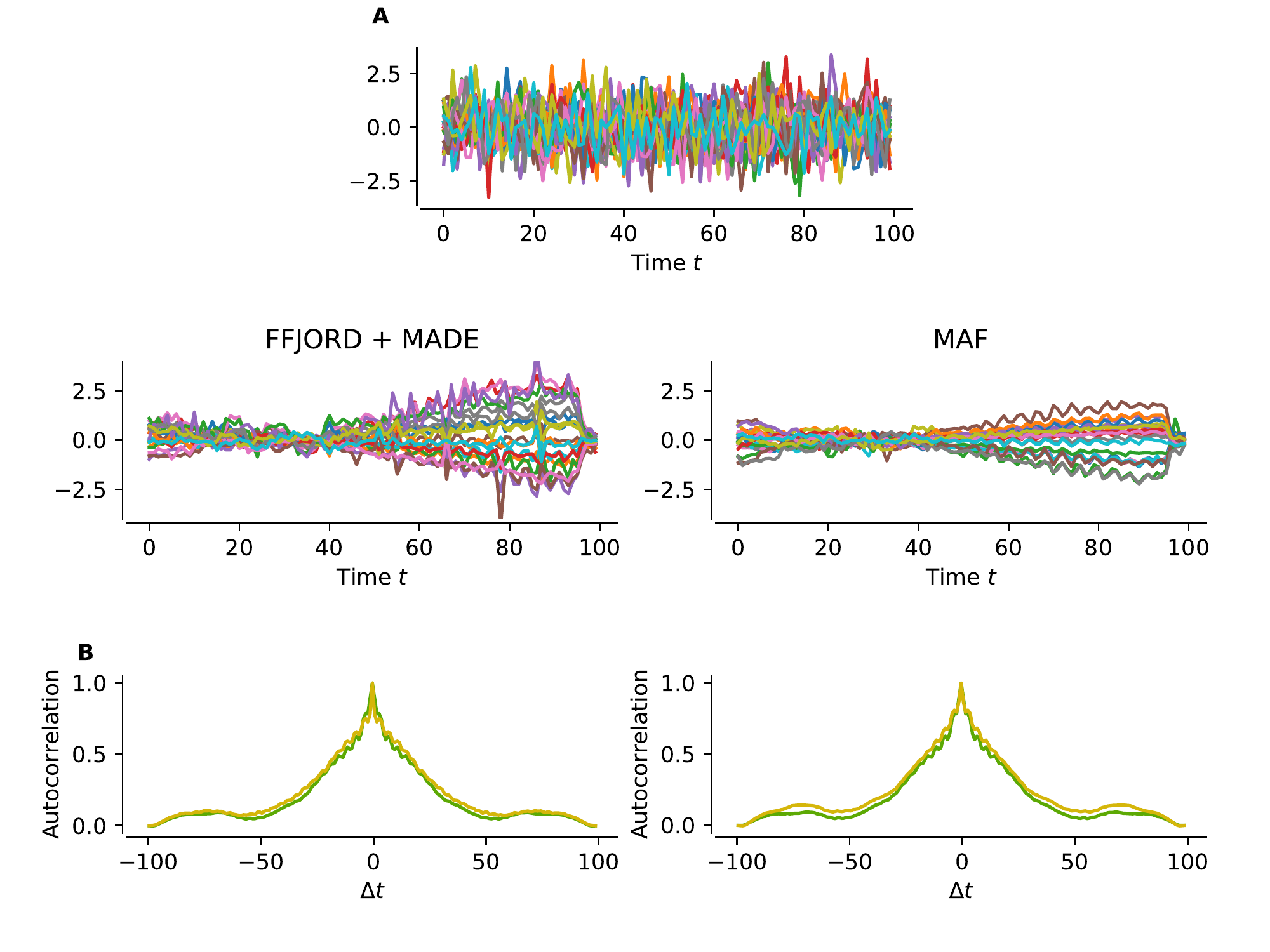}
\par\end{centering}
\caption{Generation of new data samples. (A) Samples of white noise ($z(t)\sim N(0,1)\forall t$)
(top) are transformed into samples from the data distribution with
the two trained models (bottom). (B) The autocorrelation averaged
across 20 generated samples (yellow) matches the empirical autocorrelation
of the data (green) very well. The FFJORD+MADE model was trained for
600 epochs.}
\label{fig:Fig4_generative}
\end{figure}

The flow-based models enable us to generate new, artificial data.
We draw random samples of white noise (\figref{Fig4_generative}A,
top) and pass them through the models to transform the random samples
into samples from the data distribution (\figref{Fig4_generative}A,
bottom). The generated samples fit the empirical autocorrelation very
well (\figref{Fig4_generative}B), demonstrating that we indeed can
generate artificial samples of normal data from the learned models.

\section{Discussion}

In this work, we evaluate the use of flow-based generative models
for novelty detection in time series. Since normalizing flows approximate
distribution of the data and score data samples based on their likelihood,
they possess the sufficient ingredients for novelty detection. In
comparison to other conventional methods, they can learn very flexible
distributions and allow the modeler to impose helpful constraints
such as autoregressive properties in time series.

To test our idea, we use two flow-based models: Masked Autoregressive
Flows (MAF) and a Free-form Jacobian of Reversible Dynamics (FFJORD)
continuous-time flow. We restrict the FFJORD model with autoregressive
constraint imposed through the usage of autoregressive MADE networks,
which to the best of our knowledge is the first time. We train the
models on synthetic data, which we could control to make novelty detection
challenging, and demonstrate good classification performance, outperforming
a conventional method for novelty detection, Local Outlier Factor
(LOF). Applied to less challenging real data from an industrial machine,
the models reach perfect accuracy, marginally better than LOF.

As an extension of this work, the flow-based models could be trained
to learn the transition from normal to abnormal samples. This could
support the analysis of the defect causing the emergence of abnormal
data in the industrial machine. Furthermore, the learned transition
from normal to abnormal data could be applied to new motor operation
patterns and generate abnormal data, thereby helping to understand
the possible failure modes of the machine defect.

\section{Acknowledgments}

We would like to thank Jun Yamaguchi and Hitoshi Hasunuma from Kawasaki
Heavy Industries Ltd. and Shunji Goto from Ascent Robotic Inc. for
their support of this project.

\section{Supplement\label{sec:Supplement}}

For all experiments data was preprocessed in the same way: 80\% of
normal data was used for training purposes, while remaining 20\% of
normal data, and 100\% of abnormal data was used for testing purposes.
We subsample the data by the factor of 10, and extract middle 100
time points from the whole signal. We calculate the mean and standard
deviation of the training data, and use it to normalize all three
datasets.

For the flow-based models, we used the following hyperparameters:
Adam optimizer with learning rate $0.01$ and weight decay $0.001$,
ODE solver (for FFJORD+MADE) `dopri5`, batch size $100$, number of
epochs 620 (\figref{Fig3_real_data}, FFJORD+MADE), 2000 (\figref{Fig3_real_data},
MAF), 140 (\figref{Fig3_real_data}, FFJORD+MADE), 100 (\figref{Fig3_real_data},
\figref{Fig4_generative}, MAF), 600 (\figref{Fig4_generative}, FFJORD+MADE).
Our FFJORD+MADE implementation is based on the \texttt{ffjord} library
(\url{https://github.com/rtqichen/ffjord}) and the MADE implementation
in the \texttt{pyro} library \citep{bingham2018pyro}. Our MAF implementation
is based on a publicly available implementation (\url{https://github.com/ikostrikov/pytorch-flows})
using \texttt{pytorch} \citep{paszke2017automatic}.

For Local Outlier Factor, parameter \emph{MinPts} is set to 50, and
Chebyshev distance metric was used to calculate distance between data
points. We used the implementation in \texttt{scikit-learn} \citep{scikit-learn}.

\end{document}